\pdfoutput=1

\documentclass[11pt]{article}

\usepackage{acl}

\usepackage{times}
\usepackage{latexsym}
\usepackage{subcaption}

\usepackage[T1]{fontenc}

\usepackage[utf8]{inputenc}

\usepackage{microtype}
\usepackage{graphicx}
\usepackage{pgfplots}

\graphicspath{ {./} }

\pgfplotsset{width=\linewidth, compat=1.17}

\title{Taygete at SemEval-2022 Task 4: RoBERTa based models for detecting Patronising and Condescending Language}

\author{Jayant Chhillar \\ C2FO \\ Noida, India \\
  IIIT-Delhi \\
  Delhi, India \\
  \texttt{jayant17154@iiitd.ac.in} \\}

\begin{document}
\maketitle
\begin{abstract}
This work describes the development of different models to detect patronising and condescending language within extracts of news articles as part of the SemEval 2022 competition (Task-4). This work explores different models based on the pre-trained RoBERTa language model coupled with LSTM and CNN layers. The best models achieved 15$^{th}$ rank with an F1-score of 0.5924 for subtask-A and 12$^{th}$ in subtask-B with a macro-F1 score of 0.3763. 
\end{abstract}

\section{Introduction}

The use of Patronising and Condescending Language (PCL) in text or speech can affect healthy communication channels adversely. The effect of PCL on the vulnerable sections of society have been widely studied. PCL acts as a catalyst for discriminatory behaviour \citep{mendelsohn2020framework} against various vulnerable groups. It has been observed to promote exclusion and discrimination among communities and provide a conducive environment for rumour spreading and misinformation \citep{nolan2013things}. These negative effects of PCL are unaffected by the intent of the writer/speaker who might have unknowingly used PCL. These reasons provide a strong argument for developing methods that can identify and prevent unwanted use of PCL in news articles, blogs, and other pieces of text. 

In subtask-A of the Patronizing and Condescending Language Detection task at Semeval-2022 \citep{perezalmendros2022semeval}, the goal is to develop a model which takes a sample text as an input and outputs a label indicating the presence or absence of PCL. In subtask-B, the model was required to identify the correct set of PCL categories. The model takes in a sample text as an input and return seven separate outputs each indicating the presence or absence of the pre-defined seven categories. The dataset for the task was shared by the task organisers in the English language. To tackle these tasks, RoBERTa based models were developed. Different variations of the models involved the use of feed-forward layers, LSTMs, CNN and their combinations. 
For subtask-A RoBERTa with LSTM, CNN and feed-forward layers outperformed all the other variations with an F1 score of 0.5924. In subtask-B RoBERTa with feed-forward layers got the best F1 score of 0.3763 as compared to the other variations. For subtask-A, this work achieved 15th rank in the leader board and 12th rank for subtask-B details of which are discussed in section 3.2.

\begin{table*}
\centering
\begin{tabular}{lll}
\hline
\textbf{Sentence} & \textbf{Keyword} & \textbf{Label}\\

In September , Major Nottle set off on foot from Melbourne \\ to Canberra to plead for a national solution to the homeless problem .& homeless & 1 \\
\hline
10:41am - Parents of children who died must get compensation , free\\ medicine must be provided to poor families across UP : Ram Gopal Yadav& poor-families & 1 \\
\hline
Today , homeless women are still searching for the same thing .\\ A place to sleep and be safe & homeless & 0 \\
\hline
For refugees begging for new life , Christmas sentiment is a luxury most \\ of them could n't afford to expect under shadow of long-running conflicts& refugee & 0 \\
\hline
\end{tabular}
\caption{\label{sampleData}
Sample text and keyword pairs along with corresponding labels. 
}
\end{table*}

\section{Background}

Identification and analysis of PCL in text is well explored in linguistics \citep{aggarwal2012survey}, politics \citep{huckin2002critical}, sociolinguistics \citep{thapar2016exploring} and other fields. However, in NLP it is still heavily unexplored and starting to gain traction. In the past topics such as sentiment analysis \citep{feldman2013techniques}, offensive speech identification \citep{safaya-etal-2020-kuisail} and fake news identification \citep{shu2017fake} have been significantly worked upon. One major roadblock in exploring PCL in the text is the lack of well structured and labelled dataset. Recently, some new work has been developed to tackle this issue. Wang et al. \citep{wang2019talkdown}
 developed a model for identifying the condescending language in Reddit threads and also developed an annotated dataset for the same.

\subsection{Dataset}
\begin{figure}
\input{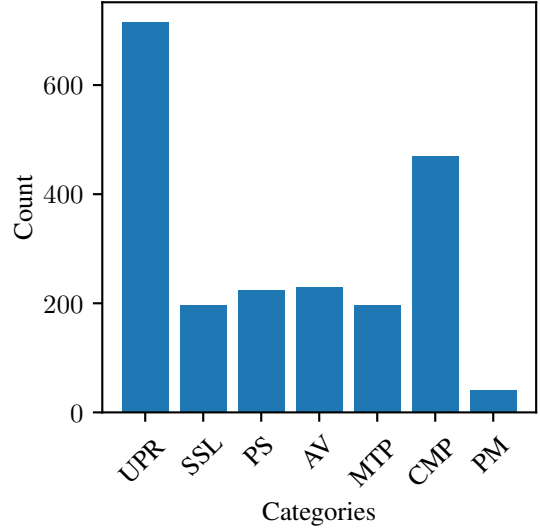}
    \caption{Number of samples for each of the seven PCL classes}
    \label{classDist}
\end{figure}

For training and development of the model presented in this work "The Don’t Patronize Me!" dataset \citep{perez-almendros2020dontpatronizeme} was used.
The dataset contains paragraphs in the English language extracted from the News on Web (NoW) corpus. It comprises 10469 samples out of which 993 have been classified as positive samples, i.e. they contain PCL. 
The dataset categorizes PCL into 7 different sub-categories, namely,
Unbalanced power relations (UPR), Shallow solution (SSL), Presupposition (PS), Authority voice (AV), Metaphor (MTP), Compassion (CMP), The poorer - the merrier (PM). Each positive sample can belong to any combination of these categories. The distribution of each category out of all the positive samples is described in Figure \ref{classDist}.

For developing the models for subtask-A the dataset also provides binary labels (0 or 1) to signify the presence or absence of PCL in the text. Along with the paragraphs, the dataset includes the country of origin of the original article and keywords that occur in the paragraph under consideration. These keywords comprise the following, Disabled, Homeless, Hopeless, Immigrant, In need, Migrant, Poor Families, Refugee, Vulnerable and Women. These keywords are usually present in texts that concern the vulnerable sections of society (refer Table \ref{sampleData}). 

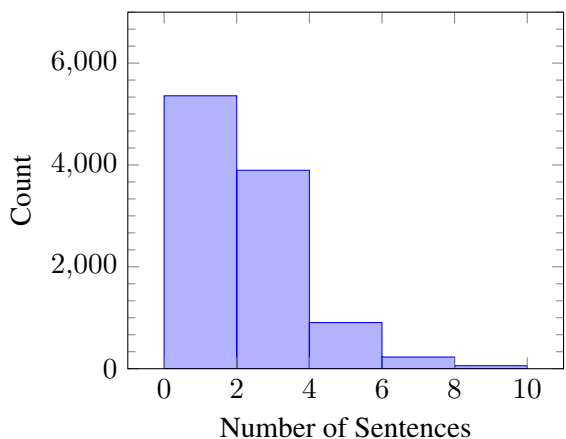
\begin{figure}
\begin{tikzpicture}
\begin{axis}[
    ymin=0, ymax=7e3,
    minor y tick num = 5,
    area style,
    ylabel=Count,
    xlabel=Number of Sentences,
    width=\linewidth*0.95
    ]
    
\addplot+[ybar interval,mark=no] plot coordinates { (0, 5358) (2, 3896) (4, 906) (6, 229) (8, 59) (10, 12) };
\end{axis}
\end{tikzpicture}
\caption{Distribution of the number of sentences per sample.}
\label{numSent}
\end{figure}

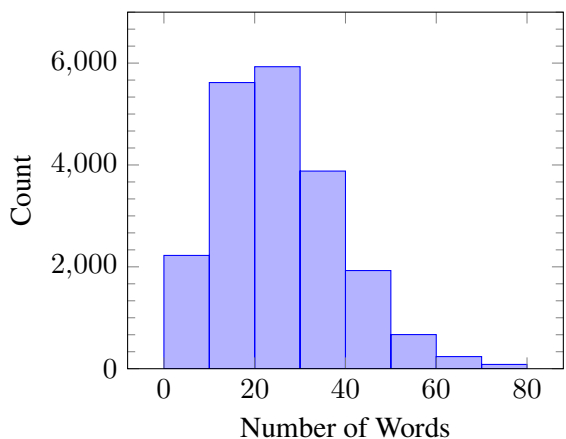
\begin{figure}
\begin{tikzpicture}
\begin{axis}[
    ymin=0, ymax=7e3,
    minor y tick num = 5,
    area style,
    ylabel=Count,
    xlabel=Number of Words,
    width=\linewidth*0.95
    ]
    
2224., 5618., 5929., 3880., 1927.,  670.,  238.,   89.
\addplot+[ybar interval,mark=no] plot coordinates { (0, 2224) (10, 5618) (20, 5929) (30, 3880) (40, 1927) (50, 670) (60, 238) (70, 84) (80, 33)};
\end{axis}
\end{tikzpicture}
\caption{Distribution of the number of words per sentence in a sample.}
\label{numWords}
\end{figure}

\section{System Overview}

This section describes the different model designs explored for Task A and Task B and the pre-processing techniques employed. Section 3.1 describes the pre-processing techniques and how they tackle the challenges offered by the dataset. The different models are described under section 3.2 along with a description of the different sub-components and the underlying intuition.

\subsection{Data pre-processing}

The "The Don’t Patronize Me!" dataset offers primarily three major challenges, which are, low number of samples, high class imbalance and the low context in the textual data (smaller sentence length). To deal with high class-imbalance and lower number of samples data augmentation techniques, loss weighting strategies were adopted and to address the low context issue, keywords shared in the dataset were used to provide added context to the models.

\subsubsection{Tokenisation}

Each sample was tokenised using $\mathrm{RoBERTa}$ \cite{liu2019roberta} tokenizer. To identify the optimal Tokenisation length analysis was done on the distribution of the number of a sentence per sample (Figure \ref{numSent}) and the distribution of the number of words for each sentence (Figure \ref{numWords}). On analysing the two distributions length of 50 to 60 tokens seemed a viable candidate for Tokenisation operation. However, on further analysis, it was found that out of 993 positive samples, 193 (19.43 \%) had more than 75 words. Thus, to prevent loss of information Tokenisation was done with a length of 100. 

\begin{figure}
  \includegraphics[width=\linewidth]{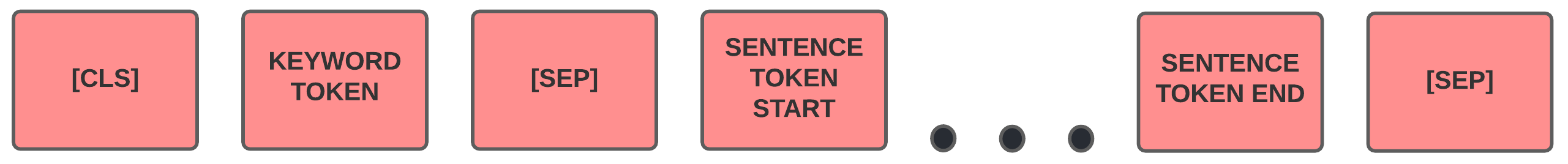}
  \caption{An example of tokenisation} 
  \label{fig:token}
\end{figure}

Each tokenised sentence was prepended with a tokenised keyword corresponding to that sample separated by the SEP token. Finally, the Tokenisation process was completed by adding a CLS and SEP token at the beginning and the end respectively (refer Figure \ref{fig:token}).  

\subsubsection{Data augmentation}

For data augmentation back-translation method was explored. Back-translation is the process of using a language model to translate a text from its parent language to another language, generally using a language model. The new text is then translated back to its parent language. This method introduces slight changes in the structures of the text while retaining the underlying context. This method has been shown to boost the performance of models trained over smaller datasets.\citep{sennrich-etal-2016-improving}.
Helsinki-NLP models \footnote{https://huggingface.co/Helsinki-NLP} were used to translate a sentence from English to French and back to English. Only 30 per cent (randomly sampled) of the positive samples from the dataset were back-translated.

\subsubsection{Loss weighting}

Initial exploratory analysis of the dataset has shown high class imbalance. To address this issue cost-sensitive re-weighting technique developed by Cui et al \cite{cui2019class} and suggested by Jurkiewicz et al \cite{jurkiewicz2020applicaai} was adopted.  The weighting factor for each class was identified as per the following definition:

\begin{equation}
(1 - \beta)/(1 - \beta^{n_i})
\end{equation}\label{eq1}

where $\beta$ is a hyper-parameter in [0,1), and $n_i$ is the number of samples belonging to the class $i$. Using these weights the updated softmax cross-entropy loss is given as:

\begin{equation}
L(z,y) = \frac{1 - \beta}{1 - \beta^{n_i}}\log{ \Bigg(\frac{\exp{(z_y)} } {\sum^{C}_{j=1}{\exp{(z_j)}} } \Bigg)}
\end{equation}\label{eq2}

where $z = [z_1, z_2, ... , z_C]$ is the predicted output of the model for $C$ classes and $y$ being one of the possible class labels, i.e. $y \in C$

\subsection{Model description}

\begin{table}
\centering
\begin{tabular}{llll}
\hline \textbf{Model} & \textbf{BASIC} & \textbf{AUG} & \textbf{WT} \\ \hline
RB-FNN & \textbf{0.6177} & 0.6301 & 0.6080 \\
RB-BiLSTM & 0.6140 & \textbf{0.6305} & 0.6258\\
RB-CNN & 0.5879 & 0.5954 & 0.6037 \\
RB-BLS-CNN & 0.6059 & 0.6095 & \textbf{0.6318}\\
\hline
\end{tabular}
\caption{\label{AUGA} Analysis of the models trained under WT, AUG and BASIC setting for subtask-A}
\end{table}

\begin{figure}
  \includegraphics[width=\linewidth]{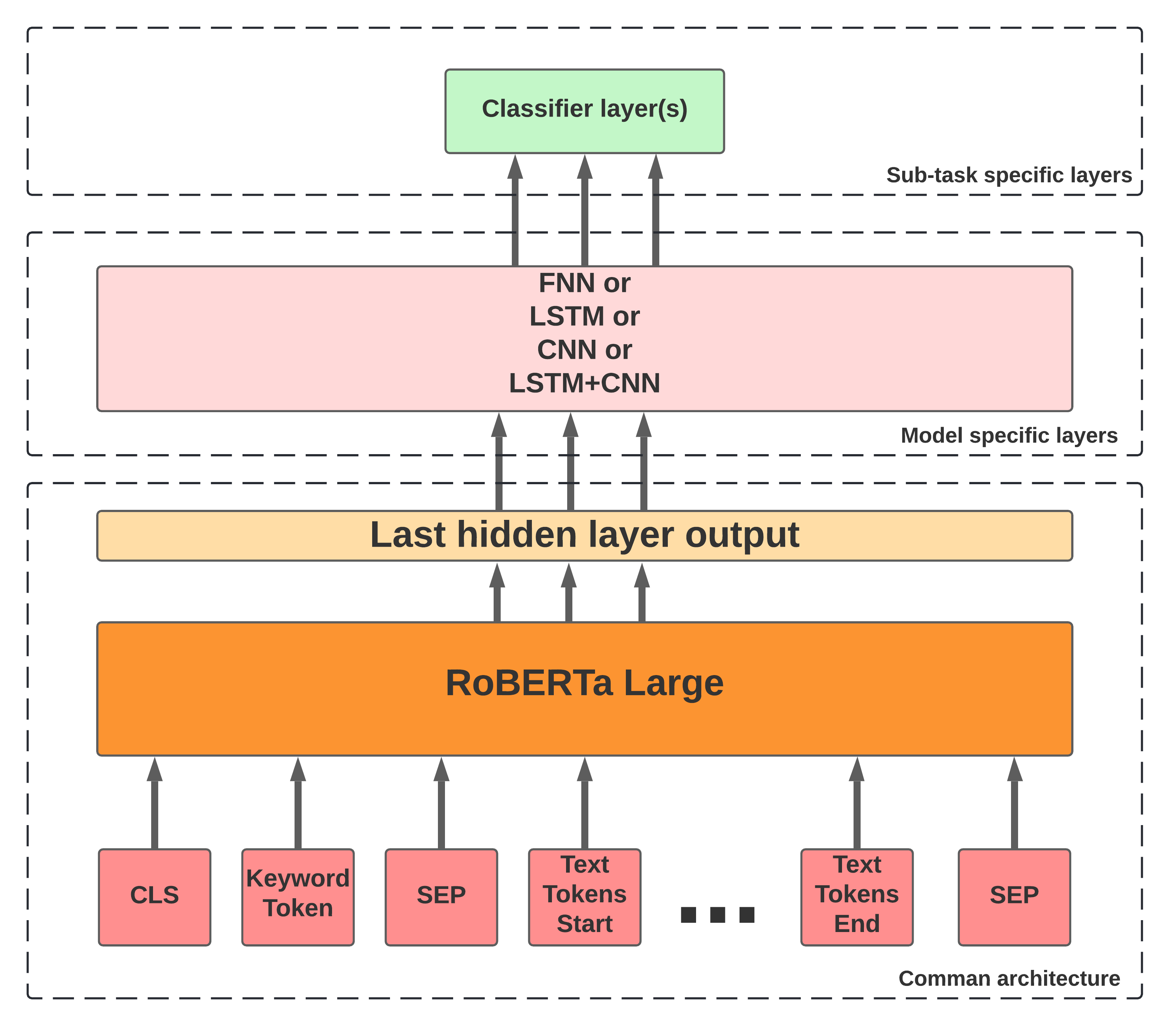}
  \caption{Generalised architecture of the models developed. For subtask-A classifier layers consist of single FNN with 2 units. For subtask-B classifier layers consist of 7 FNN layers each with 2 units.} 
  \label{fig:Gen model}
\end{figure}

This work explores four different model designs. Each design includes $\mathrm{RoBERTa}_\mathrm{LARGE}$ \cite{liu2019roberta} as it's base layer. The output of the last hidden state (shape = 106 X 1024) is then further fed down the network to get the final prediction. 
For subtask-A, all the models perform binary prediction (0 = no PCL, 1 = contains PCL) to identify if the input text contains PCL, while for subtask-B each model produces 7 binary predictions, one for each possible PCL category. The design of the four models remains the same for both tasks except for the number of outputs generated by them (Figure \ref{fig:Gen model}). Binary Cross-Entropy (BCE) loss was used for all the outputs. Adam optimizer was utilized with a learning rate set to 1e-6 and epsilon at 1e-6.

\subsubsection{RB-FNN}

The model employs the use of two feed-forward layers added on top of $\mathrm{RoBERTa}_\mathrm{LARGE}$. The output of the last hidden layer is flattened and passed down the model. The initial feed-forward layer has 106 units. For subtask-A, the output of this hidden layer is passed on to a single feed-forward layer with 2 units for binary prediction, while for subtask-B the output is shared by seven feed-forward layers each with 2 units predicting the presence of each sub-category of PCL.

\subsubsection{RB-BiLSTM}

LSTM is a type of recurrent neural network (RNN) that allows the model to learn underlying features in temporal data without the added drawbacks of general RNN models such as exploding or vanishing gradients. LSTM allows the model to capture the long term dependencies in the data and identify the underlying temporal nature of the data\citep{tang2015effective}. LSTMs have shown to achieve state of the art performance in different text classifications tasks \citep{tang2015effective} and  \citep{li2020survey}. Shi and Lin \citep{shi2019simple} also showed that using LSTM coupled with BERT can improve the performance compared to BERT by itself.
For this model the output of the last hidden layer of $\mathrm{RoBERTa}_\mathrm{LARGE}$ model is fed into a Bi-Directional LSTM layer with 106 units. The output of the BiLSTM layer is then fed down to two FNN layers with 106 and 2 units respectively (subtask-A). For subtask-B, the output of the first FNN layer is fed to seven feed-forward layers each with 2 units.

\subsubsection{RB-CNN}

CNN based models have been shown to perform well for various text classification problems \citep{chen2015convolutional} \citep{safaya-etal-2020-kuisail}. CNN layers are able to capture the semantic relationships within the textual data and given the structured nature of the embeddings obtained from $\mathrm{RoBERTa}_\mathrm{LARGE}$ model it seemed beneficial to use CNN layers to extract the hierarchical features within the data \citep{rodrigues2019multimodal}. In this model, the last layer embeddings of the $\mathrm{RoBERTa}_\mathrm{LARGE}$ model are fed to two CNN layers coupled with a max-pooling layer. The first CNN layer comprises 64 10X10 filters with stride 1 and the second layer comprises 32 5X5 filters with stride 1. After each CNN layer, a two-dimensional max-pooling operation is done with a shape of 2X2. The output of the last max-pooling operation is fed to an FNN layer with 106 units which is followed by an FNN layer with 2 units (subtask-A). For subtask-B, the output of this FNN layer is fed to seven feed-forward layers each with 2 units.

\subsubsection{RB-BLS-CNN}

To get the model to learn both temporal and hierarchical features within the data a hybrid model was developed employing both LSTM and CNN layers. This model is created as an amalgamation of the RB-BiLSTM and RB-CNN models. The last layer $\mathrm{RoBERTa}_\mathrm{LARGE}$ embeddings are fed to an LSTM layer with 106 units. The output of the LSTM layer is then further fed to the CNN architecture defined in the RB-CNN model. The final FNN layer with 106 units is then further fed to a single FNN layer with 2 units for subtask-A and to seven separate FNN layers each with 2 units for subtask-B.

\begin{table*}
\centering
\begin{tabular}{lllllllll}
\hline
\textbf{Model} & \textbf{Macro F1} & \textbf{UPR F1}& \textbf{SSL F1}&
\textbf{PS F1}& \textbf{AV F1}& \textbf{MTP F1}& \textbf{CMP F1}& \textbf{PM F1}\\

Baseline & 0.1041 & 0.3535 & 0 & 0.1667 & 0 & 0 & 0.2087 & 0\\

RB-FNN & \textbf{0.3763} & \textbf{0.5969} & \textbf{0.4578} & \textbf{0.3333} & \textbf{0.2178} & \textbf{0.3043} & \textbf{0.536} & \textbf{0.1875} \\

\hline
\end{tabular}
\caption{\label{taskB}
F1 score comparison on evaluation dataset for subtask-B between the RoBERTa baseline shared by task organisers and the RB-FNN under AUG experimental settings.
}
\end{table*}

\begin{table}
\centering
\begin{tabular}{llll}
\hline \textbf{Model} & \textbf{F1 score} & \textbf{Precision} & \textbf{Recall} \\ \hline
Baseline & 0.4911 & 0.3935 & 0.653  \\
RB-BLS-CNN & \textbf{0.5924} & \textbf{0.5357} & \textbf{0.6625} \\

\hline
\end{tabular}
\caption{\label{taskA} F1 score comparison on evaluation dataset for subtask-A between the RoBERTa baseline shared by task organisers and the RB-BLS-CNN under WT experimental settings.}
\end{table}

\section{Experimental setup}

To gauge the effect of data augmentation and loss weighting techniques on the performance of models for each subtask four experiments were carried out (Table \ref{expTable}). The goal was to identify how both the techniques interacted with each other and to find the right combination for each subtask. For each experiment, the model was trained on 80 per cent of the data as the training set and 20 per cent as the validation set. The 80-20 split shared by the task organisers was used. F1 score for subtask-A and macro F1 score for subtask-B were chosen by the task organisers as the criteria to identify the best performing model, thus the same was used to evaluate the performance of different models created for the two subtasks under different experimental settings.
For each experiment, training was done for 20 epochs with a batch size of 8. The best version of the model from each experiment was used to generate predictions for the evaluation dataset.

\begin{table}
\centering
\begin{tabular}{llll}
\hline \textbf{Model} & \textbf{BASIC} & \textbf{AUG} & \textbf{WT}  \\ \hline
RB-FNN & \textbf{0.4054} & \textbf{0.4082} & 0.3158 \\
RB-BiLSTM & 0.3643 & 0.3818 & 0.2880\\
RB-CNN & 0.3594 & 0.3903 & \textbf{0.3180}\\
RB-BLS-CNN & 0.3599 & 0.3519 & 0.2871\\
\hline
\end{tabular}
\caption{\label{AUGB} Analysis of the models trained under WT, AUG and BASIC setting for subtask-B}
\end{table}

\section{Results}

For subtask-A  RB-BLS-CNN under WT experiment achieved the highest F1 score of 0.5924 with a precision of 0.5357 and recall of 0.6625 on the evaluation dataset. While on the validation dataset the same model received an F1 score of 0.6318 with a precision of 0.5685 and recall of 0.7109. 

For subtask-B RB-FNN performed best out of all the models under the AUG experimental settings. The model achieved a macro F1 score of 0.4006 on the validation dataset and 0.3763 on the evaluation dataset.

The minute difference in the F1 scores of the best models for the evaluation dataset and the validation dataset shows that the model did not overfit during the training phase despite a large number of training epochs. 

The effect on class re-weighting (refer \ref{eq2}) and data augmentation was also explored (refer Table \ref{AUGA} and Table \ref{AUGB}). It was found that for subtask-A a majority of the four models received a boost in the F1 score when class re-weighting was applied as compared to the BASIC experimental setting. However, this trend was absent for all the models of subtask-B. Rather class-weighting had a detrimental effect on the models for subtask-B as shown in Table \ref{AUGB}. The low number of samples for each of the seven sub-classes coupled with the added complexity of the task as compared to subtask-A could have been the underlying cause behind this observation.

Similarly, the effect of data augmentation on model performance was also explored (refer to Table \ref{AUGA} and Table \ref{AUGB}). For subtask-A, all the different models received a boost as compared to the models without augmented data. The same trend persisted for the majority of models trained for subtask-B.

Another interesting find in subtask-B was the significantly poor performance of the LSTM and CNN based models as compared to the vanilla RoBERTa model i.e. RB-FNN. This is not in line with the trend observed for the models in subtask-A (Table \ref{AUGA}). The reason for this result could be similar to the unexpected trend observed for the models of subtask-B in class re-weighting experiments. Especially for LSTM based models as a large number of samples are required to train models that employ LSTMs in their design.

\begin{table}
\centering
\begin{tabular}{lcc}
\hline \textbf{Exp} & \textbf{Augment} & \textbf{Loss Weighting} \\ \hline
BASIC & No & No \\
AUG & Yes & No \\
WT & No & Yes \\
AUG+WT & Yes & Yes \\
\hline
\end{tabular}
\caption{\label{expTable} Different experiments carried out on each model. }
\end{table}

\section{Conclusion}

\begin{table*}
\centering
\begin{tabular}{llllllll}
\hline
\textbf{} & \textbf{UPR}& \textbf{SSL}&
\textbf{PS}& \textbf{AV}& \textbf{MTP}& \textbf{CMP}& \textbf{PM}\\

Precision & \textbf{0.6076} & 0.3684 & \textbf{0.5667} & \textbf{0.3428}& \textbf{0.5652}& \textbf{0.6521}& \textbf{0.6666} \\

Recall & 0.5563& \textbf{0.3889}& 0.2741& 0.3157& 0.25& 0.4245& 0.1818  \\

\hline
\end{tabular}
\caption{\label{taskB-RBFNN}
Precision and Recall values for RB-FNN model under AUG experimental settings on test data.
}
\end{table*}

This work explored the design and training of different RoBERTa based models for PCL detection in text. The added benefits of using CNN and LSTM layers along with RoBERTa in boosting model performance was also shown. This work also explored the effects of using back translation as a data augmentation technique along with a class re-weighting technique to deal with low sample size and high class imbalance. Finally, the challenges offered by the models under different problem statements were explored which gives a deeper insight into the impacts of different design methodologies. The best models achieved 15th and 12th rank for subtask-A and subtask-B respectively.

Future work can include expanding the dataset with more data as the current dataset includes 10469 samples. Also, the original article can be provided against each sample which can be fed to the model as added context. This added context should significantly boost model performance. 

\section{Acknowledgments}

The author is grateful for the support of Dr Rajiv Ratn Shah and for the guidance offered by Aradhya Mathur. The author would also like to thank the anonymous reviewers for their insightful feedback.

\bibliographystyle{acl_natbib}
\bibliography{acl_latex}


\end{document}